\documentclass[conference]{IEEEtran}
\IEEEoverridecommandlockouts

\usepackage{cite}
\usepackage{amsmath,amssymb,amsfonts}
\usepackage{algorithmic}
\usepackage{graphicx}
\usepackage{textcomp}
\usepackage{xcolor}
\def\BibTeX{{\rm B\kern-.05em{\sc i\kern-.025em b}\kern-.08em
    T\kern-.1667em\lower.7ex\hbox{E}\kern-.125emX}}
\usepackage{multirow}
\usepackage{bbding}
\usepackage{mathrsfs}
\usepackage{amssymb}
\usepackage{url}
\usepackage{array}

\begin{document}

\title{First-frame Supervised Video Polyp Segmentation via Propagative and Semantic Dual-teacher Network\\
}
\author{
Qiang Hu$^{1}$,~
Mei Liu$^{2}$,~
Qiang Li$^{1}$,~
Zhiwei Wang$^{1,\dag}$\\
\\
$^1$~Wuhan National Laboratory for Optoelectronics, Huazhong University of Science and Technology\\
$^2$~Tongji Medical College, Huazhong University of Science and Technology\\
}

\maketitle
\def\thefootnote{$\dag$}\footnotetext{Corresponding author.}

\begin{abstract}
Automatic video polyp segmentation plays a critical role in gastrointestinal cancer screening, but the cost of frame-by-frame annotations is prohibitively high.
While sparse-frame supervised methods have reduced this burden proportionately, the cost remains overwhelming for long-duration videos and large-scale datasets.
In this paper, we, for the first time, reduce the annotation cost to just a single frame per polyp video, regardless of the video's length.
To this end, we introduce a new task, First-Frame Supervised Video Polyp Segmentation (FSVPS), and propose a novel Propagative and Semantic Dual-Teacher Network (PSDNet).
Specifically, PSDNet adopts a teacher-student framework but employs two distinct types of teachers: the propagative teacher and the semantic teacher.
The propagative teacher is a universal object tracker that propagates the first-frame annotation to subsequent frames as pseudo labels. However, tracking errors may accumulate over time, gradually degrading the pseudo labels and misguiding the student model.
To address this, we introduce the semantic teacher, an exponential moving average of the student model, which produces more stable and time-invariant pseudo labels. PSDNet merges the pseudo labels from both teachers using a carefully-designed back-propagation strategy.
This strategy assesses the quality of the pseudo labels by tracking them backward to the first frame.
High-quality pseudo labels are more likely to spatially align with the first-frame annotation after this backward tracking, ensuring more accurate teacher-to-student knowledge transfer and improved segmentation performance.
Benchmarking on SUN-SEG, the largest VPS dataset, demonstrates the competitive performance of PSDNet compared to fully-supervised approaches, and its superiority over sparse-frame supervised state-of-the-arts with a minimum improvement of 4.5\% in Dice score.
Codes are at \url{https://github.com/Huster-Hq/PSDNet}.
\end{abstract}


\section{Introduction}
Colorectal cancer (CRC) is one of the most common cancers in the world, which seriously threatens to human health and life.
The development of automated polyp segmentation technologies for colonoscopy play a key role in screening and treating CRC~\cite{ji2024frontiers,yang2024dacat}.
However, most of the existing polyp segmentation methods~\cite{fan2020pranet,guo2020learn,wei2021shallow,zhou2023cross,wang2023iboxcla,hu2024monobox} are designed for colonoscopy images, ignoring the temporal information, and fail to achieve ideal performance in colonoscopy videos.
In turn, the development of video polyp segmentation (VPS)~\cite{ji2022video,fang2024embedding,hu2024sali} is limited by the shortage of dataset, as it requires frame-by-frame pixel-wise annotations, whose cost is very expensive.

To address label scarcity, there has been increasing attention on semi-supervised polyp segmentation, which leverages unlabeled data to aid model training.
Treating video frames as separate images and utilising common image-based semi-supervised segmentation methods~\cite{liu2022perturbed,wu2023acl,sun2024corrmatch} is an available solution, but they can not model the temporal correlation carried by videos.
Recently, a few works have been proposed specifically for the video domain.
For example, TCCNet~\cite{li2022tccnet} and SSTAN~\cite{zhao2022semi} introduce temporal modules to realize feature interaction between labeled frames and adjacent unlabeled frames, and then use annotations to supervise the interaction results of labeled frames.
Nevertheless, they can only perform valid supervision at the location of sparsely labeled frames, ignoring the information of unlabeled frames.
In natural scenes, IFR~\cite{zhuang2022semi} and TDC~\cite{zhuang2024infer} reconstruct the features of unlabeled frames so that the annotations of labeled frames can directly supervise the unlabeled frames.
However, they heavily rely on the consistency of the extracted frame features, which is challenging to maintain in colonoscopy videos due to frequent noise and complex inter-frame variations.

More importantly, the methods mentioned above still require sparse annotations per video, which only achieve linear scaling in annotation reduction. For instance, they might reduce the need for labeling hundreds of frames to just dozens, but the cost remains significant.
In this work, we focus on a novel task called First-Frame Semi-Supervised Video Polyp Segmentation (FSVPS), which reduces the annotation cost to the minimum: only the first frame requires manual annotation, regardless of the number of frames in the polyp video clip.

An intuitive solution of FSVPS is to utilize some universal video object segmentation (UVOS) models~\cite{cheng2022xmem,cheng2023segment,yang2023track,ravi2024sam} to produce supervision signals (i.e., pseudo labels) on unlabeled frames.
These models can automatically segment the remaining frames once a single labeled frame is provided.
However, since their working mechanism is to propagate the label along the time direction frame by frame without using semantic information, an error accumulation could happen gradually over time, leading to more and more significant degradation in the propagated labels.

In this paper, we propose a novel FSVPS method called the Propagative and Semantic Dual-Teacher Network (PSDNet). PSDNet employs a teacher-student framework with two distinct types of teachers: one for propagation and one for semantic segmentation. The idea is to adaptively merges their pseudo labels based on prediction quality. For example, near the first annotated frame, the propagative teacher's pseudo labels might be more accurate, while the semantic teacher's results become increasingly reliable over time.
Specifically, we use an off-the-shelf UVOS model as the propagative teacher, whose tracking results are used to trigger the training of the student VPS model. Simultaneously, the semantic teacher is created by applying exponential moving average (EMA) to the student model, providing more stable and time-invariant pseudo labels.
To effectively merge the results of two teachers, we introduce a back-propagation scoring method to quantitatively assess the quality of teacher-provided pseudo labels.
Specifically, inspired by the fact that high-quality pseudo labels are more likely to be spatially aligned with corresponding ground truth (GT) labels after being propagated to labeled frames, we leverage the UVOS model to propagate the pseudo label to the first frame, and consider the consistency of the propagation result with the GT label of the first frame as the quality score of the pseudo label.
Finally, the pseudo labels with higher quality scores in each unlabeled frame are selected to guide the optimization of the student model, ensuring as accurate as possible teacher-to-student knowledge transfer.


In summary, our contributions are listed as follows:
\begin{itemize}
\item~We formulate the first-frame supervised video polyp segmentation (FSVPS) task and propose a Propagative and Semantic Dual-Teacher Network (PSDNet), which employs two distinct types of teachers to jointly provide high-quality pseudo labels throughout the video by propagating and parsing semantics, respectively.
\item~We propose a back-propagation scoring method to quantitatively and accurately assess the quality of teacher-provided pseudo labels, which is used to merge the pseudo labels of two teachers.
\item~Benchmark results on SUN-SEG demonstrate that PSDNet achieves a superior performance compared to other state-of-the-arts by averagely improving $4.5\%$ in Dice on the four sub-test sets.
Specially, with approximately $1/175$ labeled data, PSDNet achieves comparable performance to the fully supervised setting.
\end{itemize}

\section{Method}
\begin{figure*}
    \centering
    \includegraphics[width=0.90\linewidth]{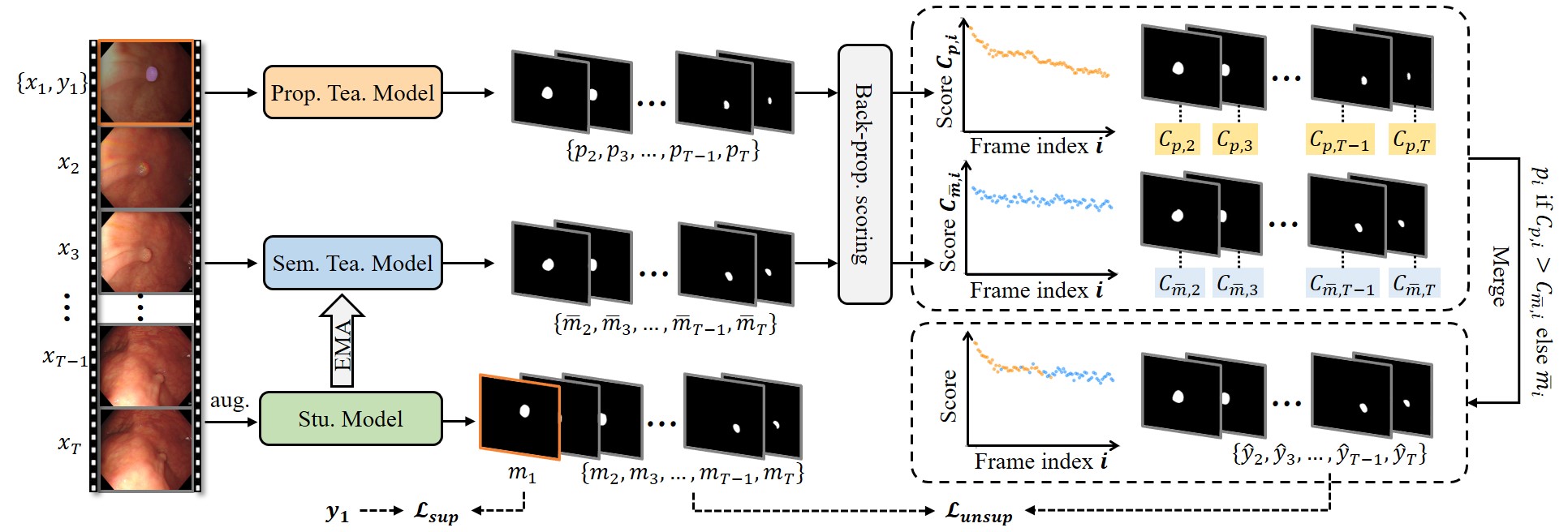}
    \caption{Overview of PSDNet. The back-propagation scoring method is used to merge the two sets of predictions of the propagative teacher model and the semantic teacher model, providing pseudo labels for the student model on unlabeled frames.}
    \label{fig:overview}
\end{figure*}

\subsection{Overview}
In this work, we foucs on the first-frame supervised video polyp segmentation (FSVPS) problem, where only the first frame of each video is annotated.
Let $\mathcal{V}=\{x_1,...,x_{T}\}$ represents a video of length of $T$ frames with $x_i \in \mathbb{R}^{3\times H \times W}$ as the $i$th frame with spatial resolution of $H \times W$, and the annotation of the whole video is $\mathcal{Y} = \{y_{1}\}$, where $y_{1} \in \{0,1\}^{H\times W}$  is a binary mask.
Assuming there are $N$ videos in the train-set, all the data we know during training is $\{ \mathcal{V}_1, ..., \mathcal{V}_N\}$ and $\{ \mathcal{Y}_1, ..., \mathcal{Y}_N\}$.
For descriptive clarity, we only consider a single video in the following part.

As shown in Fig.~\ref{fig:overview}, our proposed PSDNet adopts a teacher-student framework, which is composed of a propagative teacher, a semantic teacher and a student.
In this work, we aim to construct the final pseudo label $\hat{y}_i$ by merging the propagative teacher's output $p_i$ and the semantic model's output $\overline{m}_i$ to supervise the student's output $m_i$ on each unlabeled frames $x_i$, $i=2,...,T$.

\subsection{Two Teachers for Providing Pseudo Labels}
Considering the effectiveness and deploy-friendly, we employ XMem~\cite{cheng2022xmem} trained on natural datasets~\cite{pont20172017,xu2018youtube} as the propagative teacher model.
It propagates the annotation of the first frame and segment whole video in a class-agnostic fashion.
Specifically, we initialize its memory bank with $\{x_1, y_1\}$, then it segments the query frame by propagating the past segmented masks according to the computed correlation between the query frame and past frames. The segmentation result $p_i$ will be stored into the memory bank for segmenting the next frame, which can be formulated as:
\begin{equation}\small
    p_{i}=Prop_{1\rightarrow i}\left( Cor\left( {\left\{ {x_{1},\ldots,x_{i - 1}} \right\},x_{i}} \right),\left\{ {y_{1},\ldots,p_{i - 1}} \right\} \right),
\end{equation}
where $Prop()$ and $Cor()$ represent the function of propagation and correlation in XMem, respectively, and the subscript denotes the temporal direction of propagation.
Moreover, to better adapt the propagative teacher to colonoscopy videos, we synthesize short clips to fine-tune it by applying random affine transforms\renewcommand{\thefootnote}{\arabic{footnote}}\footnote[1]{We used rotation, sheering, zooming, translation, and cropping.} to the image-label pair of the first frame.

However, due to the error accumulation occurs in the propagation and lack of semantic knowledge about polyps, the propagative teacher cannot guarantee the long-term prediction reliability.
To address it, we create the EMA copy of the student as the semantic teacher, which is essentially a polyp semantic segmentation model.
It learns to segments each frame by encoding and decoding semantic features during training, and its segmentation result $\overline{m}_i$ is more stable and time-invariant, which is expected to be merged with $p_i$ to better supervise the student model.

\subsection{Back-propagation Scoring for Merging Pseudo Labels}
Here, we seek to merge the predictions of the two teachers based on the criteria of picking the one with higher quality.
Hence, how to accurately assess the quality of predictions is crucial.
Motivated by the fact that high-quality pseudo labels, when propagated into labeled frames, are more likely to be spatially aligned with corresponding labels, we propose a back-propagation scoring method.
Specifically, given $p_i/\overline{m}_i$, we firstly initialize XMem with $\{x_i, p_i\}/\{x_i, \overline{m}_i\}$, and let it perform back-propagation to the first frame and obtain the final results $\widetilde{p}_{1,p_i}/\widetilde{p}_{1,\overline{m}_i}$.
Next,  we take the Intersection over Union (IoU) between them and the annotation of the first frame to measure the quality score of $p_i/\overline{m}_i$, denoted as $C_{p,i}/C_{\overline{m},i}$.
Taking $C_{p,i}$ as an example, the formula is as follows:
\begin{equation}\small
    \widetilde{p}_{1,p_i}=Prop_{i \rightarrow 1}\left( Cor\left( {\left\{ {x_{2},\ldots,x_{i}} \right\},x_{1}} \right),\left\{ {\widetilde{p}_{2,p_i},\ldots,p_i} \right\} \right),
\end{equation}
\begin{equation}\small
    C_{p,i}=\frac{\vert\widetilde{p}_{1,p_i} \cap y_1\vert}{\vert\widetilde{p}_{1,p_i} \cup y_1\vert}.
\end{equation}
Finally, we choose the prediction with higher quality score in $p_i$ and $\overline{m}_i$ as the final pseudo label $\hat{y}_i$ for the student model.

\subsection{Loss Functions}
We use the annotation $y_1$ and constructed pseudo labels $\{\hat{y}_2,\ldots \hat{y}_T\}$ to guide the student model on labeled and unlabeled frames, respectively, by minimizing the standard cross-entropy (CE) loss.
The student model is optimized by minimizing the overall loss, which can be formulated as:
\begin{equation}\small
\mathcal{L}_{total} = \mathcal{L}_{sup} + \mathcal{L}_{unsup} = \mathcal{L}_{CE}(m_1,y_1) + \sum_{i = 2}^{T}{\mathcal{L}_{CE}(m_i,\hat{y}_i)}.
\end{equation}
The semantic teacher model's weights are EMA updated by the student model's weights.

\section{Experiments}
\subsection{Datasets and Evaluation Metrics}
We conduct experiments on SUN-SEG~\cite{ji2022video}, the largest polyp video segmentation dataset.
The train set contains $112$ video clips with a total of $19,544$ frames, and the test set includes four sub-test sets, SUN-SEG-Seen-Easy ($33$ clips/$4,719$ frames), SUN-SEG-Seen-Hard ($17$ clips/$3,882$ frames), SUN-SEG-Unseen-Easy ($86$ clips/$12,351$ frames), and SUN-SEG-Unseen-Hard ($37$ clips/$8,640$ frames).
Easy/Hard indicates that difficult levels to be segmented of the samples, and Seen/Unseen indicates whether the clips are sampled from the same video as train set.
In this work, only the GT mask for the first frame of each video clip in the train set is provided, with approximately $175(\frac{19544}{112})\times$ annotation mitigation.

For comprehensive comparison, we employ three metrics to evaluate the segmentation results, including Dice, intersection over union (IoU) and mean absolution error (MAE).

\subsection{Implementation Details}
\label{sec:Implementationdetetails}
Our proposed method is implemented with the PyTorch~\cite{paszke2019pytorch} framework on a single NVIDIA GeForce RTX 4090 GPU with 24GB memory.
We use SALI~\cite{hu2024sali} with PVT~\cite{wang2022pvt} pre-trained on ImageNet~\cite{deng2009imagenet} as the semantic teacher model and student model, and the weight of EMA is set to $0.999$.
We reisze the input images into $352\times352$, and set the batch size to $14$.
We use Adamw~\cite{kingma2014adam} optimizer for training with a weight decay of $0.001$.
We train the model for $30$ epochs and set the learning rate to $1e-4$.

\subsection{Comparisons with State-of-the-art Methods}
\begin{table*}[t]
    \centering
    \caption{ Quantitative comparisons with state-of-the-art semi-supervsied methods. The best performance is marked in bold.}
   {
    \begin{tabular}{lccccccccccccccccccccccccc}
    \hline
    \multirow{2}{*}{Methods} &\multirow{2}{*}{Modality} &\multicolumn{3}{c}{Seen-Easy}  & \multicolumn{3}{c}{Seen-Hard}& \multicolumn{3}{c}{Unseen-Easy}  & \multicolumn{3}{c}{Unseen-Hard}\\
		& &Dice~$\uparrow$ &IoU~$\uparrow$ &MAE~$\downarrow$ &Dice~$\uparrow$ &IoU~$\uparrow$ &MAE~$\downarrow$&Dice~$\uparrow$ &IoU~$\uparrow$ &MAE~$\downarrow$&Dice~$\uparrow$ &IoU~$\uparrow$ &MAE~$\downarrow$\\
        \hline
            UB~ &\multirow{2}{*}{-} &0.927 &0.875 &0.011 &0.891 &0.827  &0.022 &0.825 &0.751 &0.030 &0.822 &0.748 &0.027 \\
            LB~ &&0.757 &0.658 &0.043 &0.673 &0.562 &0.057 &0.607 &0.496 &0.071 &0.603&0.492 &0.068\\
            \hline
            PSMT~\cite{liu2022perturbed} &\multirow{4}{*}{Image} &0.828 &0.764 &0.027 &0.761 &0.686 &0.047 &0.660 &0.580 &0.051 &0.681 &0.606 &0.043 \\
            ACL-Net~\cite{wu2023acl} &&0.827 &0.760 &0.027 &0.757 &0.676 &0.052 &0.666 &0.589 &0.049 &0.687 &0.612 &0.046\\
            PCMT~\cite{xia2024novel} &&0.844 &0.782 &0.021 &0.788 &0.716 &0.044 &0.687 &0.614 &0.041 &0.704 &0.629 &0.039\\
            CorrMatch~\cite{sun2024corrmatch}&&0.856 &0.799 &0.019 &0.813 &0.733 &0.036 &0.733 &0.641 &0.048 &0.725 &0.630 &0.046\\
            \hline
            IFR~\cite{zhuang2022semi} &\multirow{4}{*}{Video} &0.851 &0.791&0.020 &0.808 &0.729 &0.036 &0.723 &0.636 &0.049 &0.715 &0.624 &0.048\\
            SSTAN~\cite{zhao2022semi} &&0.860 &0.799 &0.019 &0.812 &0.731 &0.036 &0.733 &0.644 &0.047 &0.717 &0.622 &0.045\\
            TDC~\cite{zhuang2024infer} &&0.863 &0.801 &0.019 &0.817 &0.738 &0.036 &0.732 &0.642 &0.047 &0.736 &0.641 &0.045\\
            TCCNet~\cite{li2022tccnet} &&0.866 &0.805 &0.019 &0.822 &0.742 &0.035 &0.751 &0.660 &0.044 &0.744 &0.652 &0.042\\
            PSDNet (Ours)~ &&\textbf{0.900} &\textbf{0.831} &\textbf{0.016} &\textbf{0.860} &\textbf{0.787} &\textbf{0.027} &\textbf{0.798} &\textbf{0.719} &\textbf{0.035} &\textbf{0.806} &\textbf{0.724} &\textbf{0.030}\\
        \hline
    \end{tabular}
    \label{table1}
    }
\end{table*}

\begin{figure*}
    \centering
    \includegraphics[width=0.9\linewidth]{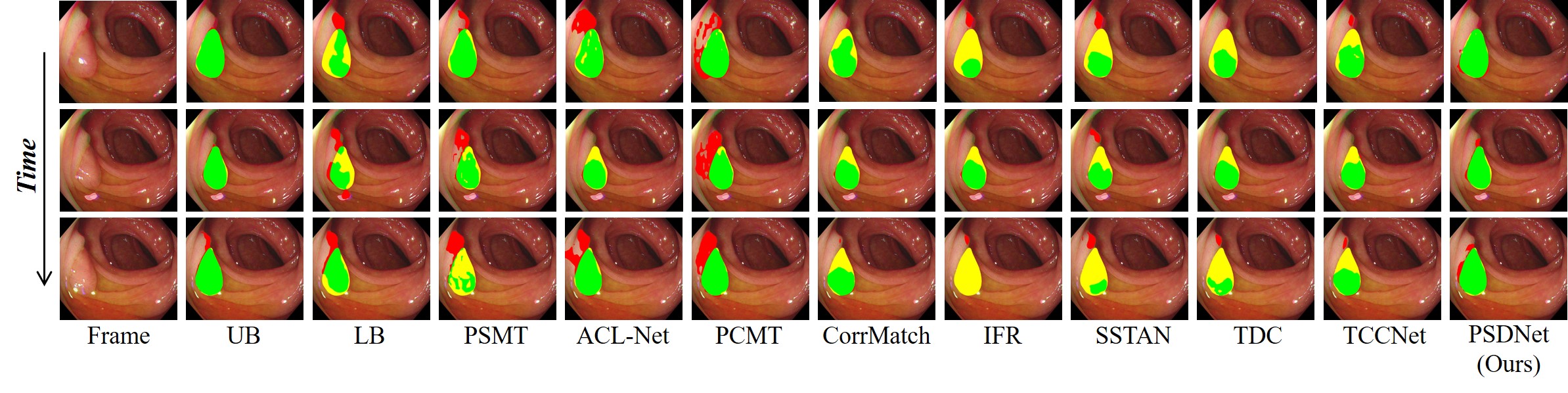}
    \vspace{-0.4cm}
    \caption{Visualization results of different methods on a challenge clip. Green: true positives; Yellow: false negatives; Red: false positives.}
    \label{fig:visualization results}
\end{figure*}

We compare PSDNet with eight semi-supervised state-of-the-arts (SOTAs), including four image-based methods and four video-based methods.
To ensure fairness, we implement all competitors through the same baseline segmentation network (i.e., SALI) with pvt as the backbone and keep the same training setting as in Sec.~\ref{sec:Implementationdetetails}.
Note that, we denote UB and LB as SALI trained on the fully-supervised setting and only the first frame, respectively.

Table~\ref{table1} presents the quantitative comparison results.
From it, we can see that annotation reduction leads to a large performance gap between LB and UB.
All of these semi-supervised methods can bring varying levels of performance improvement, and video-based methods generally outperforms image-based methods because they can additionally use temporal information.
Moreover, PSDNet achieves the best performance on all metrics of the four sub-test sets.
Specifically, it surpasses the second-best method, i.e., TCCNet~\cite{li2022tccnet}, by 3.4\%, 3.8\%, 4.7\%, and 6.2\% in Dice on four sub-test sets, respectively.
Importantly, with approximately $1/175$ labeled data, PSDNet only lags behind the UB by 2.7\%, 3.1\%, 2.7\%, and 1.6\% in Dice on four sub-test sets, respectively.

Fig.~\ref{fig:visualization results} shows the visualization results of different methods on a challenge video clip.
It can be seen that PSDNet can obtain more accurate segmentation results, as well as better maintain temporal consistency of the predictions, and overcome the motion blur and light noise in the colonoscopy video.

\subsection{Ablation Study}
\label{sec:ablation study}
\begin{table}[t]
    \centering
    \vspace{-0.2cm}
    \caption{Ablation studies of different components, including the propagative teacher model \textbf{PT}, fine-tuning the propagtive teacher with synthetic clip \textbf{FT}, the semantic teacher model \textbf{ST}, and back-propagation scoring \textbf{BS}.}
   {
    \begin{tabular}{ccccccc}
    \hline
 \multicolumn{4}{c}{Components}  &\multirow{2}{*}{Teacher}&Train set &Test set\\
 PT &FT &ST &BS & &Dice~$\uparrow$ &Dice~$\uparrow$\\
    \hline
    \XSolidBrush &\XSolidBrush &\XSolidBrush &\XSolidBrush &None &- &0.660\\
    \hline
    \Checkmark &\XSolidBrush &\XSolidBrush &\XSolidBrush &\multirow{3}{*}{Single}  &0.652 &0.687 \\
    \Checkmark &\Checkmark &\XSolidBrush &\XSolidBrush & &0.814 &0.765 \\
    \XSolidBrush &\XSolidBrush &\Checkmark &\XSolidBrush & &0.756 &0.701 \\
    \hline
    \Checkmark &\Checkmark &\Checkmark &\XSolidBrush &\multirow{2}{*}{Dual} &0.821 &0.774 \\
    \Checkmark &\Checkmark &\Checkmark &\Checkmark & &\textbf{0.893} &\textbf{0.841} \\
    \hline
    \end{tabular}
    \label{table2}
    }
    \vspace{-0.2cm}
\end{table}
We conduct a series of ablations studies to verify the effectiveness of our proposed components in Table~\ref{table2}.
Besides the average Dice on four sub-test sets, we report the Dice of all methods on the train set, which can represent the quality of pseudo labels.
Compared to the baseline, adding either the propagative or the semantic teacher model can improve model performance.
When using FT, the quality of pseudo labels provided by the propagative teacher model improves from 65.2\% to 81.4\% in Dice, and the model performance also improves from 68.7\% to 76.5\%.

If directly employ the dual-teacher, the student model receives pseudo labels from two teacher models in parallel, getting limited performance improvement (77.4\% \emph{vs.} 76.5\%) compared to the single-teacher, this is because low-quality pseudo labels are not eliminated.
With back-propagation scoring added, we obtain significant improvements of 7.2\% and 6.7\% in Dice on the train set and test set, respectively, and achieve the best performance, which shows that back-propagation scoring can effectively select high-quality pseudo labels and optimize the training of the framework.

\subsection{Comparisons with Universal Video Segmentation Methods}
\begin{table}[t]
    \centering
    \vspace{-0.2cm}
    \caption{Comparison results with prompt-based universal video segmentation methods.}
   {
    \begin{tabular}{p{1.05cm}p{0.57cm}<{\centering}p{0.57cm}<{\centering}p{0.57cm}<{\centering}p{0.57cm}<{\centering}p{0.57cm}<{\centering}p{0.57cm}<{\centering}c}
    \hline
    Method &\multicolumn{2}{c}{SAM-Track} &\multicolumn{2}{c}{SAM~2} &\multicolumn{2}{c}{MedSAM-2} &PSDNet\\
    Prompt &LB &GT &LB &GT &LB &GT &-\\
    \hline
    Dice~$\uparrow$ &\multirow{2}{*}{0.515} &\multirow{2}{*}{0.697}  &\multirow{2}{*}{0.619}  &\multirow{2}{*}{0.763}  &\multirow{2}{*}{0.643} &\multirow{2}{*}{0.771} &\multirow{2}{*}{\textbf{0.841}} \\
    (Test set) & & &\\
    \hline
    \end{tabular}
    \label{table3}
    }
    \vspace{-0.2cm}
\end{table}
Universal video object segmentation (UVOS) has attracted increased attention due to its strong zero-shot capacity, which can segment any object of interest in videos driven by user prompts.
Since it can address label scarcity in another paradigm, we compare PSDNet with UVOS SOTAs, including SAM-Track~\cite{cheng2023segment}, SAM~2~\cite{ma2024segment}, and MedSAM-2~\cite{zhu2024medical}.
We implement two formats of prompts at the first frame in each video, one is the segmentation result of LB in Table~\ref{table1}, and the other is the GT mask, denoted as LB and GT, respectively.
The results are presented in Tabel~\ref{table3}, PSDNet exceeds the second-best method by 7.0\% in Dice, which shows that training-based method still has advantages over current UVOS models in a specific task.

\section{Conclusions}
In this work, we proposed PSDNet, a novel efficient framework for FSVOS, which can greatly reduce labeling cost in VPS.
We use a UVOS model to assist, and propose a semantic teacher and back-propagation scoring to supplement semantic information and evaluate the quality of pseudo-labels, respectively.
Experiments on SUN-SEG demonstrate that PSDNet outperforms other SOTAs significantly and gains competitive performance compared to the fully supervised setting.
The framework is also expected to solve other medical video segmentation tasks with scarce data, showing great potential in clinical applications.

\section*{Acknowledgment}
This work was supported in part by National Key R\&D  Program of China (Grant No. 2023YFC2414900), Key R\&D Program of Hubei Province of China (No.2023BCB003), Wuhan United Imaging Healthcare Surgical Technology Co., Ltd.

{
    \bibliographystyle{IEEEtran}
    \bibliography{reference}
}


\end{document}